\documentclass[11pt,letterpaper]{article}
\usepackage{cogsys}
\usepackage[T1]{fontenc}
\usepackage{times}
\usepackage[pdftex]{graphicx} 

\usepackage{natbib}
\usepackage{bm}
\usepackage{multirow}
\cogsysheading{X}{20XX}{1-6}{X/20XX}{X/20XX}

\ShortHeadings{Concept Formation for Masked Word Prediction}
              {X.\ Lian, N.\ Baglodi, and C.\ J.\ MacLellan}

\begin{document} 

\title{Incremental and Data-Efficient Concept Formation to Support Masked Word Prediction}
 
\author{Xin Lian}{xinlian@gatech.edu}
\author{Nishant Baglodi}{nbaglodi3@gatech.edu}
\author{Christopher J.\ MacLellan}{cmaclell@gatech.edu}
\address{Teachable AI Lab, Georgia Institute of Technology, Atlanta, GA 30332 USA}
\vskip 0.2in
 
\begin{abstract}
This paper introduces {\it Cobweb/4L}, a novel approach for efficient language model learning that supports masked word prediction. The approach builds on Cobweb, an incremental system that learns a hierarchy of probabilistic concepts. Each concept stores the frequencies of words that appear in instances tagged with that concept label. 
The system utilizes an attribute-value representation to encode words and their surrounding context into instances. 
Cobweb/4L uses the information-theoretic variant of category utility and a new performance mechanism that leverages multiple concepts to generate predictions. 
We demonstrate that with these extensions it significantly outperforms prior Cobweb performance mechanisms that use only a single node to generate predictions.
Further, we demonstrate that Cobweb/4L learns rapidly and achieves performance comparable to and even superior to Word2Vec. 
Next, we show that Cobweb/4L and Word2Vec outperform BERT in the same task with less training data.
Finally, we discuss future work to make our conclusions more robust and inclusive.

\end{abstract}

\section{Introduction}


Over the past decade, there has been significant advancement in language models that utilize neural network architectures, particularly in tasks such as word prediction and text completion. The concept of the {\it cloze procedure}, introduced by \cite{taylor1953cloze}, measures communication effectiveness, and in particular, it is a method of ``intercepting a message from a `transmitter', mutilating its language patterns by deleting parts, and so administering it to `receivers' that their attempts to make the patterns whole again potentially yield a considerable number of {\it cloze} units'' \citep[pp. 416]{taylor1953cloze}. This idea inspired \cite{devlin-etal-2019-bert} to create the {\it MLM (masked language modeling)} paradigm, which is the task used during the training of Bidirectional Encoder Representations from Transformers ({\it BERT}). In this paradigm, certain input tokens are randomly masked, and the model predicts them solely based on the words in their surrounding contexts. Specifically, the architecture employs a bidirectional transformer encoder to predict masked tokens. This approach is widely adopted to train large language models, particularly in derivatives of BERT \citep{liu2019roberta, baevski2020advances}. Other approaches share a similar rationale. For example, {\it autoregressive} language modeling uses it in their training process; they predict the next token based on previously predicted results \citep{yang2019xlnet, brown2020language}.

Apart from MLM, the exploration into how neural-network-based architectures predict words within their contexts dates back much earlier. For example, the {\it Word2Vec} \citep{mikolov2013efficient} variant known as Continuous Bag-of-Words ({\it CBOW}) learns word embeddings by training a neural language model to predict a masked word given the context words around it.

Despite the rapid development and superior performance of such neural language models for word-filling tasks, they have several shortcomings. The prevalent approach for utilizing large language models involves training followed by fine-tuning. The training phase demands substantial data and computational resources. For example, BERT's training corpus comprised the {\it BookCorpus} (800M words) \citep{zhu2015aligning} and {\it English Wikipedia} (2.5B words) \citep{devlin-etal-2019-bert}. Additionally, it is questionable if large language models, like BERT, can learn continually.
While various approaches to continual or online learning have been proposed in recent years \citep{loureiro-etal-2022-timelms, jin-etal-2022-lifelong-pretraining, hu2023meta}, with either online training or fine-tuning, more work is needed to develop approaches that let large language models learn continuously.



Some studies diverge from the mainstream focus on large language models and instead explore more {\it human-like} learning systems to address their limitations. For instance, \cite{mitchell2018never}'s {\it NELL (Never-Ending Language Learner)} employs an infinite loop akin to an Expectation Maximization (EM) algorithm for continual learning. Additionally, \cite{maclellan2022efficient} proposes a novel approach utilizing the {\it Cobweb} algorithm \citep{fisher1987knowledge}, an approach for learning probabilistic concept hierarchies that was inspired by human cognition, for language induction. This approach incrementally builds modular structures based on prior knowledge, enabling rapid expertise acquisition from limited training data, aligning with the human-like learning constraints suggested by \cite{langley2022computational}. Although the use of Cobweb in language induction is still in its nascent stages and its practicality and performance on certain language tasks remain uncertain, it represents an initial exploration into leveraging human-like learning mechanisms into language tasks. 



%

This paper makes the following contributions:
\begin{itemize}
\item We introduce {\it Cobweb/4L}, which inherits the Cobweb structure and learns language from instances that describe words and their surrounding context, as well as a new performance mechanism for categorizing test items to generate predictions; 
\item We show that Cobweb/4L's performance mechanism yields substantially better predictive performance on a masked word prediction task than prior Cobweb performance mechanisms;
\item We present a preliminary evaluation comparing Cobweb/4L to Word2Vec, showing that it achieves better masked word prediction performance and is more data efficient;
\item We present a preliminary evaluation showing Cobweb/4L can achieve better masked work prediction performance than BERT, even when BERT is trained on three times as much data.
\end{itemize}

\noindent While these results are still preliminary, they demonstrate the potential for Cobweb-based language modeling approaches to be competitive with mainstream neural language models.


\section{Cobweb as a Language Learning System}

\subsection{Background}
{\it Cobweb} \citep{fisher1987knowledge} was first proposed as an incremental approach for learning hierarchical concepts. Given a sequence of instances represented in nominal attribute-value pairs (e.g. \textit{\{color: blue; form: triangle; number: 2\}}), Cobweb forms a probabilistic concept hierarchy, where each concept stores the attribute value count frequencies for the features of instances it has incorporated.

\begin{figure}[t!]
\vskip 0.05in
\begin{center}
\includegraphics[width=4.2in]{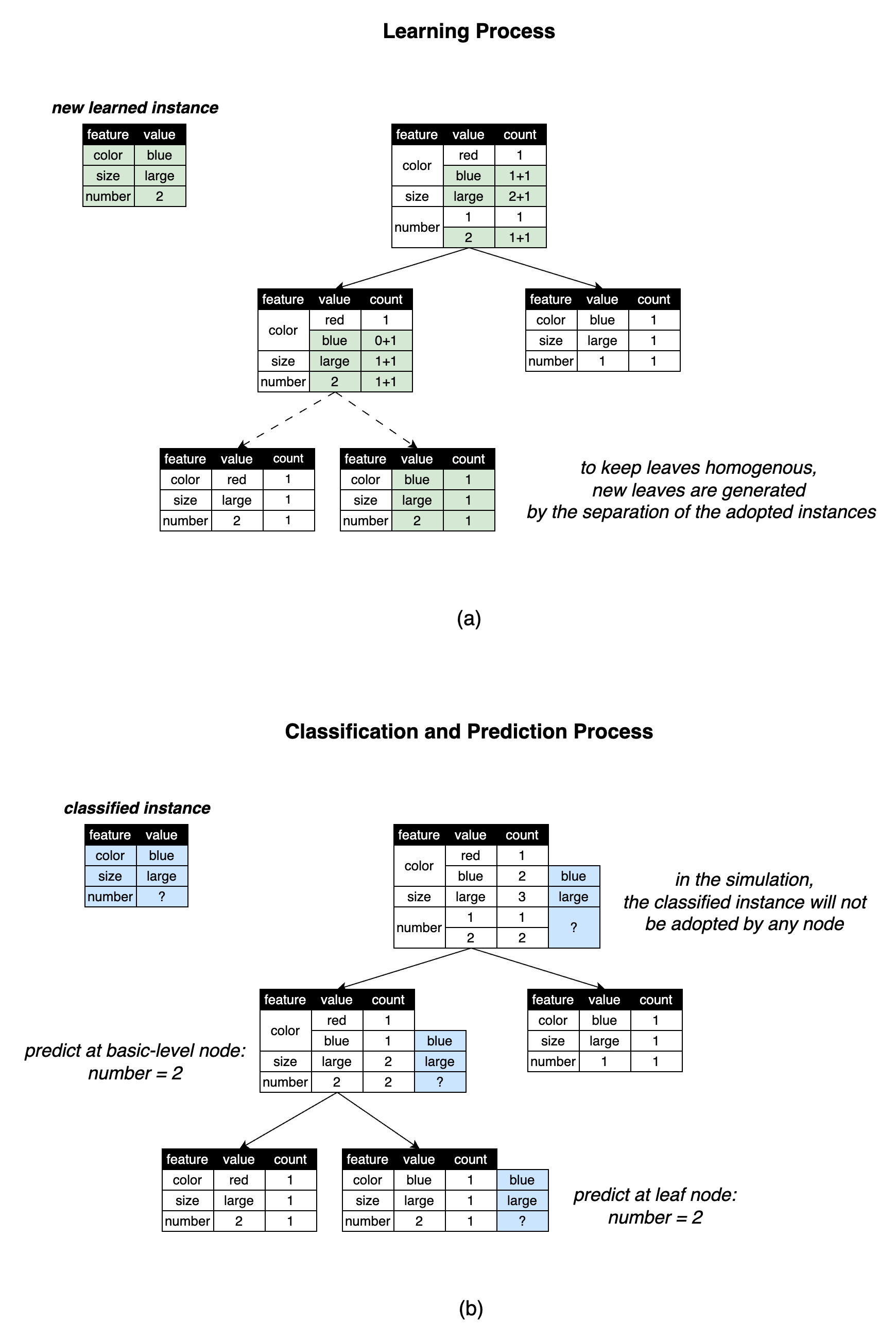}
\caption{The illustration of the learning and prediction processes of Cobweb. Figure (a) shows how Cobweb incorporates a new instance into its tree from its root to a leaf node, and Figure (b) shows how Cobweb classifies an instance with its unobserved attribute ({\tt number}) from its root to a leaf node with a similar process as during learning. To predict the unobserved attribute, Cobweb uses a specific node along the categorization path, and the two most prevalent choices are the leaf and basic-level nodes.} 
\label{fig:cobweb-tr-te}
\end{center}
\vskip -0.2in
\end{figure}

The learning process of Cobweb is illustrated in Figure~\ref{fig:cobweb-tr-te}(a). To guide the concept formation process, Cobweb uses \textit{Category Utility} \citep{corter1992explaining}, which evaluates the resulting increase in the predictive power of a concept $c$:
\begin{equation}
    CU(c) = P(c)[\sum_i \sum_j P(A_i = V_{ij}|c)^2 - \sum_i \sum_j P(A_i = V_{ij})^2]
\end{equation}
$P(c)$ is the overall probability of the concept $c$, $P(A_i = V_{ij})$ is the probability of some instance having attribute-value  $A_i = V_{ij}$, and $P(A_i = V_{ij}|c)$ is the conditional probability of some instance having attribute-value $A_i = V_{ij}$ given its membership in $c$.

When a new instance is introduced to be learned, Cobweb sorts it down its current categorization tree, and at each node $c$, it considers a {\it best} way to incorporate the instance into the current node's children from four available operations: {\it add, create, merge, split}. It selects the option that produces the highest averaged category utility among the children of $c$. This recursive process continues until the instance is located at some leaf node. 

After constructing the Cobweb tree, it can apply its structure to classify another instance having some unobserved attribute(s) and then predict them, which is illustrated in Figure~\ref{fig:cobweb-tr-te}(b). In particular, Cobweb first sorts the instance down from its root to some leaf node just like during learning. During categorization, none of the concept frequencies are updated and only the {\it add} operation is considered. Consequently, the tree will not be updated to reflect the instance. After classification, Cobweb predicts the unobserved attribute(s) of the instance using one of the nodes along the categorization path \citep{lian2024cobweb}. One option is to predict with the {\it basic-level} node, suggested by \cite{corter1992explaining} because the node holds the highest category utility value. Another option is to directly predict with the {\it leaf} node, which has been shown to yield better performance in prior work \citep{maclellan2016trestle, maclellan2022convolutional, maclellan2022efficient}.

Cobweb was not explored as a language-learning system until \cite{maclellan2022efficient} proposed three extensions to process words and their surrounding contexts---the {\it word}, {\it leaf}, and {\it path} variants. The first combines ideas from both Word2Vec techniques within the framework of probabilistic concept hierarchies, and the other variants build on this initial scheme and apply more sophisticated ways to incorporate information about word contexts during both the predicting and learning process. However, there is not a direct comparison between it and Word2Vec, or other language models, regarding performance or data efficiency.

\subsection{Cobweb/4L: A New Representation and Performance Mechanism for Improved Language Modeling}

\begin{figure}[t!]
\vskip 0.05in
\begin{center}
\includegraphics[width=3in]{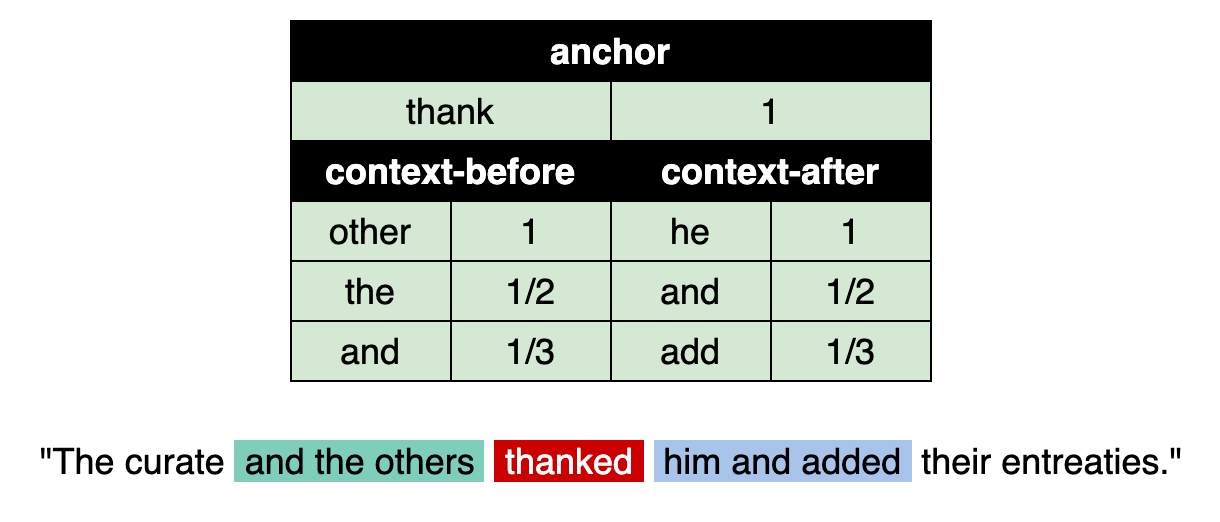}
\caption{An example of a Cobweb/4L instance derived from the text ``{\it The curate and the others thanked him and added their entreaties.}'' Suppose we want to predict the anchor word {\it thank(ed)} given the context window 3. The anchor word and the context words before ({\it and the others}) and after ({\it him and added}) are stored in the instance. Each word is stored with a weighted count that reflects the relative position from the anchor. The word has more weight if it is closer to the anchor.
}
\label{fig:inst-rep}
\end{center}
\vskip -0.2in
\end{figure}

Building on Cobweb, we propose {\it Cobweb/4L}. It utilizes an instance representation that is similar to the one used by the {\it Word} variant proposed by \citet{maclellan2022efficient}. As shown in Figure~\ref{fig:inst-rep}, given a sentence, a chosen anchor word $w_a$, and the window size, we generate an instance with attributes {\tt anchor}, {\tt context-before} and {\tt context-after}, so the values are mapped from the words in the sentence. Specifically, {\tt anchor} denotes the anchor word considered, and {\tt context-before} and {\tt context-after} attributes contain the context words before and after the anchor word in the sentence. For each context word $w$ in either attribute considering the context in the instance, instead of assigning it with the actual count in the context, we assign it with a weighted count $1/(d+1)$ where $d$ is the number of words (or tokens) between $w$ and $w_a$, so the closer $w$ gets to $w_a$, the more weight it gets, and consequently its weighted count reflects the relative position within the instance.

\begin{figure}[t!]
\vskip 0.05in
\begin{center}
\includegraphics[width=5in]{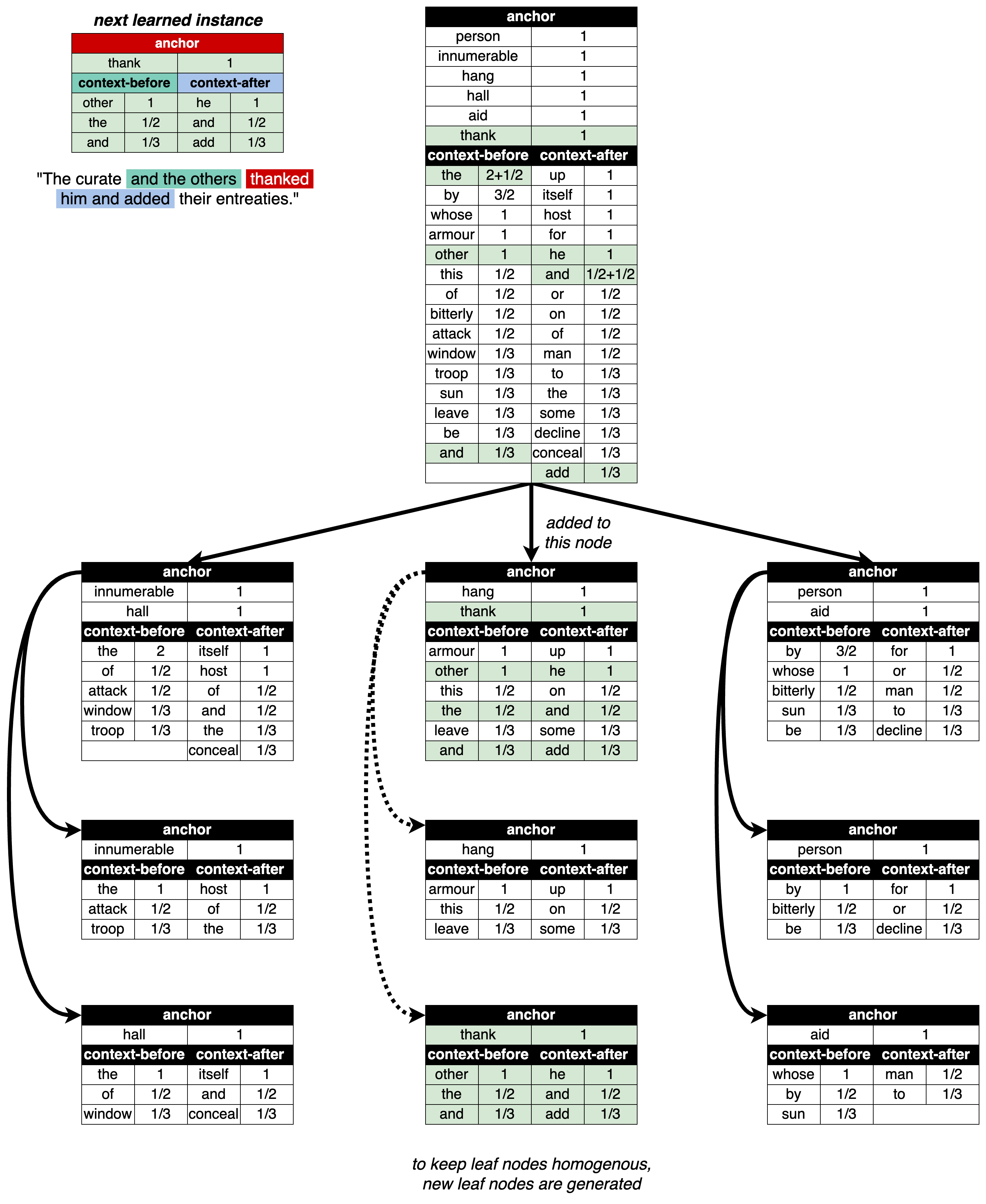}
\caption{The process of how Cobweb/4L learns an additional new instance after learning 6 instances. The process is indeed the same as the one in Cobweb (but here in particular, the information-theoretic category utility is used for evaluating the operation for each traversed concept node).
}
\label{fig:cobweb-tr}
\end{center}
\vskip -0.2in
\end{figure}

After parsing and preprocessing the raw texts into the instance format represented in Figure~\ref{fig:inst-rep}, Cobweb/4L learns using an approach similar to Cobweb, which is illustrated in Figure~\ref{fig:cobweb-tr}. Unlike prior Cobweb work, we make use of the {\it information-theoretic category utility} \citep{corter1992explaining} when deciding the operation to proceed, so the category utility of a concept node $c$ becomes
\begin{equation}
    CU(c) = P(c)[U(c_p) - U(c)]
\end{equation}
where $c_p$ is the parent of $c$ and $U(c)$ is the {\it uncertainty} (or {\it entropy}) of $c$:
\begin{equation}
    U(c) = \sum_i P(W_i|c) U(W_i|c) 
\end{equation}
where
\begin{equation}
    U(W_i|c) = -\sum_j P(w_{ij}|c)\log P(w_{ij}|c)
\end{equation}
is the uncertainty of the attribute ({\tt anchor}, {\tt context-before}, {\tt context-after}) $W_i$ given $c$, and $P(w_{ij}|c)$ is the probability that $W_i$ has word $w_{ij}$ given $c$.

\begin{figure}[t!]
\vskip 0.05in
\begin{center}
\includegraphics[width=6in]{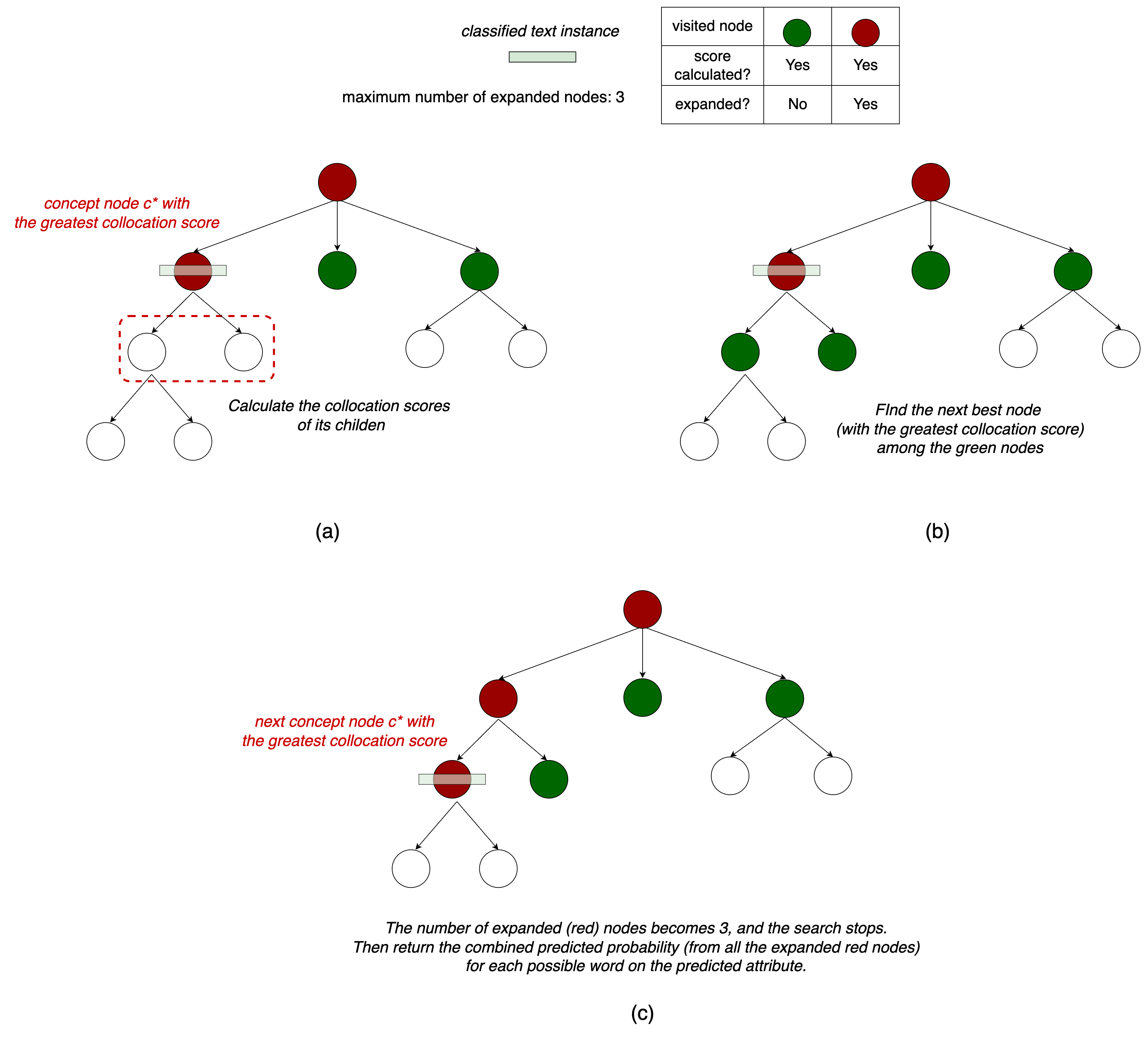}
\caption{The illustration of how Cobweb/4L classifies and predicts the word of an unobserved attribute of an instance. Starting from the root, Cobweb/4L recursively finds the concept node that has the greatest collocation score among the search frontier (green nodes), and after finding the best node $c^*$, it adds $c^*$ to the collection of expanded nodes (red nodes), extends the search frontier to its children (so its children turn green), and continues to find the next best concept node, until the number of expanded nodes reaches the number of maximum expanded nodes $N_{max}=3$. After that, Cobweb/4L calculates the predicted probability for a certain word under the unobserved attribute by combining the predicted probabilities from all expanded nodes weighted by their collocation scores.
}
\label{fig:cobweb-pred}
\end{center}
\vskip -0.2in
\end{figure}

Cobweb/4L uses a new performance mechanism that leverages multiple nodes within the tree to generate predictions. 
Given a maximum number of nodes $N_{max}$ to expand, it begins categorization at the root. Instead of sorting the classified instance $\bm{x}$ down a single, greedy path as is typically done in with Cobweb, it utilizes a kind of {\it best-first search}. It maintains a {\it search frontier} and expands the {\it best} concept node $c^*$ at each step.
The best node is the one that has the greatest {\it collocation} score \citep{jones1983identifying}:
\begin{equation}
    s(c) = P(c|\bm{x})P(\bm{x}|c)
\end{equation}
which is the product of cue and category validity. Once a new $c^*$ has been identified, Cobweb/4L adds it to the expanded node list $\mathcal{C}^*$, adds its children to the search frontier, and repeats.
This process starts at the root $c^* = c_{root}$ and repeats recursively by evaluating the collocation of the children of $c^*$ and adding these to the search frontier to be considered for expansion (see Figure~\ref{fig:cobweb-pred}). The process continues until Cobweb/4L has expanded $|\mathcal{C}^*| = N_{max}$ nodes. After that, Cobweb/4L predicts the word $w_i$ for the attribute $W_i$ with probability:
\begin{equation}
    P(W_i = w_i | \mathcal{C}^*) = \sum_{c\in\mathcal{C}^*}P(w_i|c)\frac{\exp\{-s(c)\}}{\sum_{c\in\mathcal{C}^*}\exp\{-s(c)\}}
\end{equation}
which represents the combination of predictions from all expanded nodes, weighted by their collocation (the softmax ensures the probabilities sum to one). This approach is a kind of Bayesian model averaging \citep{hinne2020conceptual}.

\section{Experiments}

In our experiments, we conducted a comprehensive performance comparison between Cobweb/4L and two other baseline architectures, CBOW and BERT, with the datasets used in the \textit{Microsoft Research (MSR) Sentence Completion Challenge} \citep{zweig2011microsoft}.

\subsection{Compared Baseline Architectures}

\subsubsection{Word2Vec and CBOW}
Word2Vec \citep{mikolov2013efficient} was initially proposed as two feedforward neural network architectures for computing continuous vector representations from a given corpus. One is CBOW, which is trained to predict the anchor word given the context words around it. The order of words does not influence the projection from words to vectors (i.e., they are a bag of words). The other is {\it Continuous Skip-gram}, which is similar to CBOW, but predicts surrounding words given the anchor word. \cite{mikolov2013efficient} compared the Skip-gram model with other language models, such as 4-gram and average LSA (latent semantic analysis) on the MSR sentence completion challenge task. They found that the combination of Skip-gram and recurrent neural network language model works the best among all objectives being compared. Our experiment here compared Cobweb/4L with CBOW in the word completion task given context. In particular, we trained CBOW with 3 epochs, batch size 64, and vector length 100.

\subsubsection{Transformers and BERT}

The BERT architecture \citep{devlin-etal-2019-bert} is a multi-layer bidirectional Transformer encoder \cite{vaswani2017attention}. In training the two initial versions of BERT ({\tt BERT-base} and {\tt BERT-large}), \cite{devlin-etal-2019-bert} used two tasks: one is the MLM, where they randomly substituted {\tt [MASK]} for some tokens in the training data\footnote{In their original training process, they randomly chose some tokens to be manipulated, then substituted 80\% of the selected tokens with {\tt [MASK]}, 10\% with some other random token, and left the remaining 10\% unchanged.}. The other is the {\it Next Sentence Prediction}, where each sentence is paired with another sentence. Each pair is annotated with whether the sentences occurred together in the training data or not. In our work, we only train BERT with the MLM task using the architecture available at {\it Hugging Face} with 1 epoch and batch size of 64.

\subsection{Methodology}
To evaluate an approach to the MSR sentence completion challenge task, each architecture is trained with the collection of 522 Conan Doyle's \textit{Sherlock Holmes} novels. We first preprocessed the stories with lemmatization and tokenization to generate the training set for each evaluated architecture from these novels. We filtered out tokens that appeared less than three times. Next, we generated the corresponding shuffled (with fixed random seed 123) input representations from these tokens for each evaluated architecture. In particular, for Cobweb/4L, we generated attribute-value instances with a window size of 10---each instance comprises one anchor word, 10 context words before, and 10 context words after. For CBOW, each trained datum comprises one anchor word and its corresponding set of context words (also with a window size of 10). For BERT, the training data is just the collection of tokenized texts.

To compare the performance and the data efficiency between Cobweb/4L and CBOW, after generating the respective preprocessed training dataset for each evaluated architecture, we trained each with approximately one-third of the data. This corresponds to the 5 million training training examples for Cobweb/4L and CBOW (each of which has the same anchor and context words).\footnote{We intended to train each baseline with all available text data, but we only had the time and resources to train on a portion of the data.} We saved a checkpoint for each model at 12 points throughout training---Cobweb/4L and CBOW were incrementally trained on 416667 examples for each checkpoint.
The training data provided between each checkpoint was approximately $1/36$th of all the Sherlock Holmes stories ($1/3 \times 1/12 = 1/36$).

To evaluate each checkpoint, we applied it to predict a single masked word in each of the 1040 test sentences. Each sentence comes with a list of five possible words for the masked slot. The words were chosen to be reasonable options that have similar frequencies in the overall corpus. Here is one example:

\begin{quote}
    {\it I have it from the same source that you are both an orphan and a bachelor and are \_\_\_\_ alone in London.
    \\A. crying\ \ B. instantaneously\ \ \textbf{C. residing}\ \ D. matched\ \ E. walking}
\end{quote}

\noindent Each baseline generated probabilities for all possible masked words and we selected the option that has the highest probability. We then compared the selected word with the ground-truth answers to calculate accuracy. 

We evaluated the architectures under this paradigm because this task, as a sentence completion (or masked word predicting) task, has a prevalent use in evaluating Word2Vec \citep{mikolov2013efficient} and other similar language models that consider context \citep{melamud2016context2vec}. For each approach, we tested each checkpoint to observe their performance throughout training and to derive their learning curves, so we can figure out which approach is more data efficient and reaches asymptotic performance faster.

\begin{figure}[t!]
\vskip 0.05in
\begin{center}
\includegraphics[width=5.5in]{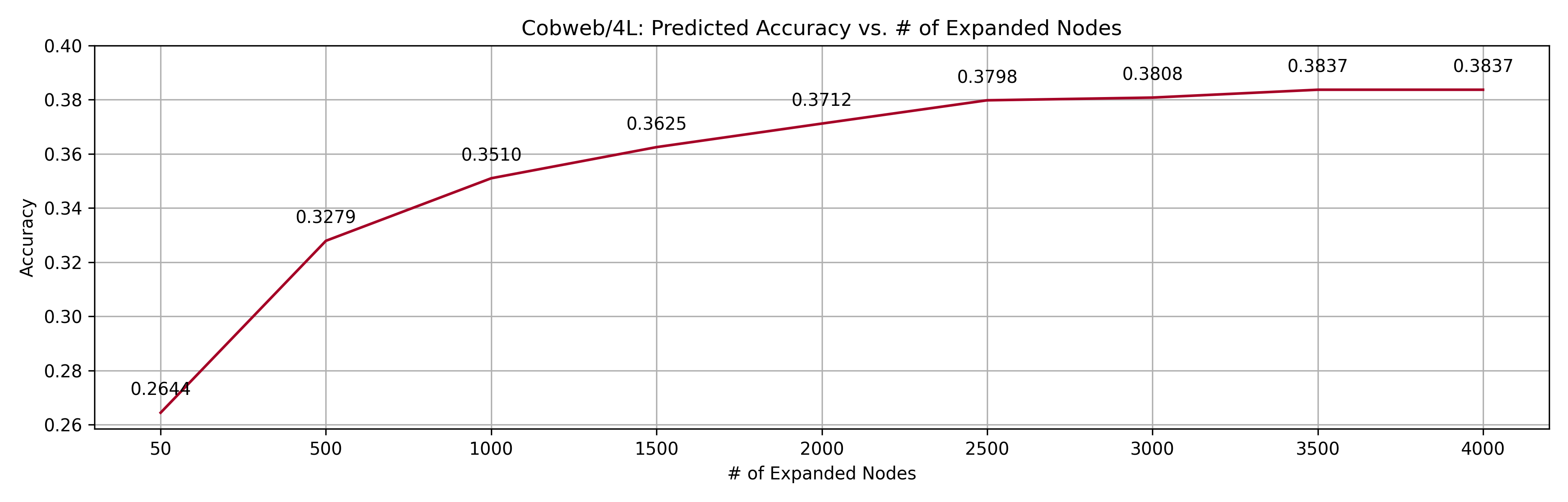}
\caption{Cobweb/4L's prediction accuracy on the MSR Sentence Completion Challenge \citep{zweig2011microsoft} data after training approximately one-third of the Sherlock Holmes Stories data with varying maximum number of expanded nodes in prediction $N_{max}$.
}
\label{fig:cobweb-tr-varying}
\end{center}
\vskip -0.2in
\end{figure}

Note that after training all the Cobweb/4L checkpoints, and before testing each of them, we additionally implemented a preliminary experiment to evaluate how performance changes based on the maximum expanded nodes $N_{max}$ during prediction. In particular, we tested Cobweb/4L (at its last checkpoint, so it was trained with one-third of the training data) with varying $N_{max}$ on the same test items. Our results, shown in Figure~\ref{fig:cobweb-tr-varying}, indicate that expanding more nodes yields higher accuracy, though it needs more computation time, and the performance plausibly levels off at $N_{max}=4000$. 
For subsequent experiments, we decided to try three options for $N_{max}$: $1000$, $2000$, and $3000$. When evaluating Cobweb/4L at each checkpoint, we compare its multi-node prediction with its {\it single-node} prediction methods (at either leaf or basic-level node) in parallel, so that we can determine if the new multi-node prediction method outperforms the single-node ones.


After evaluating Cobweb/4L and CBOW, we also evaluate BERT in a similar fashion. However, instead of training on just one-third of the data, we trained BERT with {\it all} the data. We compared its test accuracy at the last checkpoint and its peak test accuracy with those generated by Cobweb/4L and CBOW (both of which trained on one-third of the data). We want to see if they could outperform BERT with much less training data.

In our hypothesis, although the basic-level prediction might perform better for its closer alignment to human learning \citep{lian2024cobweb}, we expected Cobweb/4L, using its multi-node prediction approach, to outperform significantly the two prior Cobweb single-node prediction methods. This is because when either the leaf or basic-level node is chosen for prediction, Cobweb/4L restricts its consideration solely to the words stored within the chosen nodes, and these limited prediction choices are generally detrimental to word prediction. In contrast, Cobweb/4L's multi-node prediction strategy starts from the root node, so it considers all encountered words in each prediction. In addition, its best-first search prioritizes expanded nodes containing words pertinent to the contextual information.
Further, when more nodes are expanded during prediction, we expect Cobweb/4L to achieve higher accuracy over the course of training and to potentially have a steeper rate of learning.

When compared to other neural network architectures, we anticipate Cobweb/4L to perform similar to or better than CBOW, which is known to outperform several other approaches on this task \citep{mikolov2013efficient}. While BERT's performance in this specific task remains uncertain, it is typically used and evaluated in more complex tasks. Since smaller models trained on domain-specific data may exhibit better performance within their respective domains compared to larger models trained on a diverse range of data, it would not be surprising if Word2Vec and/or Cobweb/4L outperform BERT on this task.

\subsection{Results and Discussion}

\begin{figure}[t!]
\vskip 0.05in
\begin{center}
\includegraphics[width=6in]{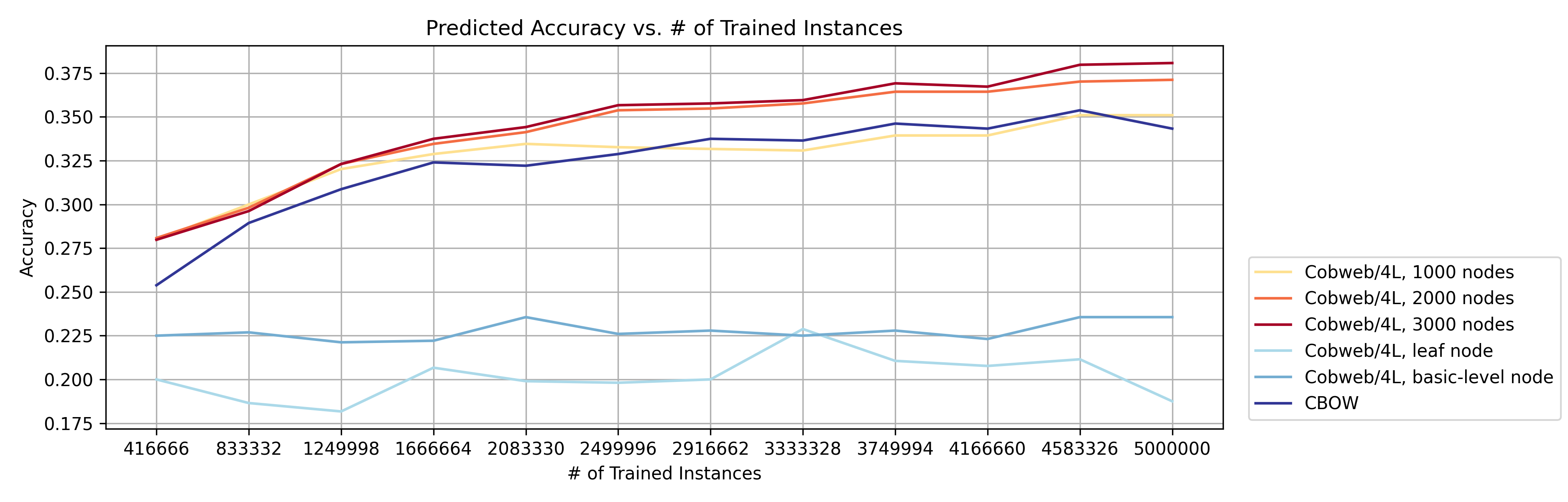}
\caption{The accuracy for each evaluated approach on the MSR sentence completion challenge test items at each of the 12 checkpoints.
}
\label{fig:cobweb-result}
\end{center}
\vskip -0.2in
\end{figure} 

The test accuracy results on the challenge data across the 12 checkpoints for each evaluated architecture are shown in Figure~\ref{fig:cobweb-result}. The single-node prediction approaches (leaf and basic-level nodes) used in prior Cobweb work do not perform well on this language modeling task. They display relatively unstable and inferior test accuracy across the checkpoints. Despite the increased number of training instances processed at each checkpoint, the learning curves for both single-node prediction methods fail to demonstrate improvement. Specifically, the basic-level prediction maintains an accuracy of approximately 0.225, while the leaf prediction hovers around 0.2 with notable variance (essentially chance performance). Conversely, Cobweb/4L's new multi-node prediction method consistently outperforms both single-node prediction approaches. In particular, all three learning curves with varying expanded nodes start with a test accuracy of about 0.28, surpassing the best performance for the single-node approaches at the outset. They exhibit a general trend of improvement at each successive checkpoint. Additionally, we observe that when more nodes are expanded, we observe more rapid accuracy improvements. 


Compared to CBOW, Cobweb/4L (with its multi-node prediction) shows more rapid learning, with higher performance during the earlier checkpoints. At later checkpoints, the 1000-node version of Cobweb/4L achieves similar performance to CBOW, while the 2000- and 3000-node versions exceed CBOW's performance.

\begin{table}[t!]
\vskip -0.15in
\caption{Test accuracies on MSR Sentence Completion Challenge testing data at either the last checkpoint or the respective peak for all evaluated architectures, including Cobweb/4L (with multi-node prediction, \{1000, 2000, 3000\} expanded nodes), Cobweb/4L (with either leaf or basic-level node prediction), CBOW, and BERT. Note that we trained Cobweb/4L and CBOW on merely one-third of the Sherlock Holmes stories, and we (pre)trained BERT on all the stories.}
\label{tab:compare}
\small
\begin{center}
\begin{tabular}{c||ccccc|c|c}
\hline\hline
\multirow{2}{*}{{\bf Accuracy at}} & \multicolumn{5}{c|}{{\bf Cobweb/4L}} & \multirow{2}{*}{{\bf CBOW}} & \multirow{2}{*}{{\bf BERT}}\\
 & 1000 & 2000 & 3000 & Leaf & Basic-level & & \\ \hline
 Last & 0.3510 & 0.3712 & 0.3808 & 0.1875 & 0.2356 & 0.3433 & 0.3125 \\
 Peak & 0.3510 & 0.3712 & 0.3808 & 0.2288 & 0.2356 & 0.3538 & 0.3125 \\
\hline\hline
\end{tabular}
\end{center}
\vskip -0.10in
\end{table}

Further, Table~\ref{tab:compare} shows the test accuracy for all evaluated approaches at their respective last checkpoint and their peaks. Although we trained BERT with all the available training data, its test accuracy still cannot surpass the accuracy achieved by Cobweb/4L with multi-node prediction and CBOW.

The results are consistent with all the aforementioned hypotheses, and we show that Cobweb/4L, with its newly introduced multi-node prediction approach, can learn more rapidly and achieve higher accuracy than the other approaches, including CBOW, which has been shown to outperform several other techniques on this task.


\section{Related Work}
While few systems operate under a rationale similar to Cobweb, we here mention several notable alternatives. A {\it decision tree (DT)} employs a {\it top-down} induction process, beginning with a root and extending down to its leaves, with a similar flowchart- (or tree-) like structure of Cobweb. However, in a DT, each internal node represents a logical test ({\it split}) and each leaf constitutes a prediction \citep{costa2023recent}. One DT approach, {\it CART (Classification and Regression Tree)} \citep{breiman1984classification}, learns from instances consisting of attributes and a target attribute in a greedy manner, iteratively selecting the most ``informative'' splits. Its procedure continues until all leaves contain pure conditions, so it potentially results in overfitting, which can be mitigated by tree pruning. Later studies introduced non-greedy trees in general, such as {\it lookahead} trees \citep{norton1989generating, ragavan1993lookahead}, which selects the optimal splits relating to the next $k$ iterations with a $k$-lookahead. Another non-greedy DT method involves building multiple trees, randomly modifying and combining them using ``genetic operators'' until a satisfactory solution is achieved, known as {\it Evolutionary Algorithms (EA)} \citep{mitchell1998introduction}. Various alternative DT induction approaches have been extensively explored, including {\it gradient-based} optimization \citep{jordan1994hierarchical, suarez1999globally}, recently more popularized by the artificial neural network community \citep{costa2023recent}, and {\it optimal} trees which search for the tree that maximizes some measure under size constraints (e.g. maximum depth) \cite{costa2023recent}. Most initial DT approaches are non-incremental, requiring the entire dataset upfront, such as CART and {\it ID3} \citep{quinlan1986induction}. Later, alternative approaches supporting incremental learning with a tree emerged, including {\it ID4} \cite{schlimmer1986case}, EA, Cobweb, {\it ID5R} \citep{utgoff1989incremental, kalles1996efficient}, etc.

Beyond DT approaches, other human-like learning systems operate similarly, such as {\it SAGE (Sequential Analogical Generalization Engine)} proposed by \cite{mclure2010learning}. SAGE maintains a generalization context for each concept, featuring a ``trigger'' to assess whether an incoming example should be adopted. With each new example, it utilizes MAC/FAC to retrieve up to three examples or generalizations based on similarity to the new example. It also learns concepts incrementally, and its later variations, equipped with the {\it Nearest-Merge} algorithm \citep{liang2014constructing}, organize concepts hierarchically. We mention these approaches as we believe they should also be explored as potential alternatives to neural language modeling, similar to how we have been exploring extensions to Cobweb for this purpose. For example, we are interested in the possibility of employing decision tree learning or other systems like SAGE for language model learning.

\section{Conclusion and Future Work}

In this paper, we introduced Cobweb/4L, a novel language modeling system based on Cobweb. It employs a new representation and a new multi-node performance mechanism that leverages several concepts in its hierarchy to generate predictions. We show this new approach significantly outperforms prior Cobweb performance mechanisms that only use single-node to generate a prediction. Further, we show the new approach learns more rapidly and achieves better accuracy than CBOW or BERT. While our results are preliminary, they suggest that a Cobweb-based language modeling approach has the potential to be both data efficient and accurate.

Moving forward, our future work entails devoting more time and computational resources to evaluating Cobweb/4L and CBOW checkpoints that are trained on the remaining available instances generated from all the Holmes stories. Additionally, to ensure robust experimental results and draw more conclusive findings, we plan to conduct the same experiment with multiple random seeds. Furthermore, we will explore variations of the multi-node prediction method and compare the information-theoretic category utility used by Cobweb/4L with the probabilistic category utility measure employed in prior Cobweb research. In more extensive studies, we envision evaluating Cobweb/4L across a broader range of language modeling tasks, including other masked word prediction and autoregressive text generation.

\begin{acknowledgements} 
\noindent
This research was partially funded by Award 2112532 from NSF’s AI-ALOE institute and Awards W911NF2120101, W911NF2320203, and W911NF2120126 from ARL’s STRONG program. The views, opinions, and findings expressed are the authors' and should not be taken as representing the official views or policies of these funding agencies. Additionally, we extend our sincere appreciation to individuals who generously shared their invaluable insights and feedback on this work, notably Pat Langley (ISLE), Jesse Roberts (Vanderbilt), Kyle Moore (Vanderbilt), and Doug Fisher (Vanderbilt).
\end{acknowledgements} 


{\parindent -10pt\leftskip 10pt\noindent
\bibliographystyle{cogsysapa}
\bibliography{format}

}


\end{document}